\documentclass{article}

\usepackage{graphicx} 
\usepackage{amsmath,amssymb,amsthm}
\usepackage{bm}
\usepackage{algorithm}
\usepackage{algorithmic}

\usepackage{mathtools}
\usepackage{bbm} 
\usepackage{comment}
\usepackage{tikz}
\usepackage{enumitem}
\usepackage{subcaption}
\usepackage{float}
\usepackage{booktabs}
\usepackage{graphics}
\usepackage{setspace}
\usepackage{siunitx}
\usepackage[
    colorlinks=false,
    linkbordercolor=green,
    citebordercolor=green,
    pdfborder={0 0 1} 
]{hyperref}
\usepackage{authblk}
\usepackage[sorting=none]{biblatex} 
\title{Graph Regularized PCA}

\author[1, 3]{Antonio Briola\thanks{Corresponding author: \href{mailto:antonio.briola@shell.com}{a.briola@ucl.ac.uk}}\thanks{The authors declare no competing interest.}} 
\author[3]{Marwin Schmidt}
\author[3, 4]{Fabio Caccioli}
\author[1]{Carlos Ros Perez}
\author[1]{James Singleton}
\author[2]{Christian Michler}
\author[3, 4]{Tomaso Aste}

\affil[1]{Shell Information Technology Limited, London, SE1 7NA, UK}
\affil[2]{Shell Information Technology International B.V., The Hague, 2596 HR, Netherlands}
\affil[3]{University College London, Department of Computer Science, London, WC1E 6EA, UK}
\affil[4]{Systemic Risk Centre, London School of Economics and Political Sciences, London, WC2A 2AE, UK}

\begin{document}

\date{}
\maketitle

\subsection{Abstract}
Multivariate data often exhibit complex dependencies that violate the assumption of
isotropic residual noise. For such cases, we introduce \emph{Graph Regularized PCA} (GR-PCA). It is a graph-based regularization of PCA that incorporates the dependency structure of the data features by learning a sparse precision graph and biasing loadings toward the low-frequency Fourier modes of the corresponding graph Laplacian. Consequently, high-frequency signals are suppressed, while graph-coherent low-frequency ones are preserved, yielding interpretable principal components aligned with conditional relationships. We evaluate GR-PCA on synthetic data spanning diverse graph topologies, signal-to-noise ratios, and sparsity levels. Compared to mainstream alternatives, it concentrates variance on the intended support, produces loadings with lower graph-Laplacian energy, and remains competitive in out-of-sample reconstruction. When high-frequency signals are present, the graph Laplacian penalty prevents overfitting, reducing the reconstruction accuracy but improving structural fidelity. The advantage over PCA is most pronounced when high-frequency signals are graph-correlated, whereas PCA remains competitive when such signals are nearly rotationally invariant. The procedure is simple to implement, modular with respect to the precision estimator, and scalable, providing a practical route to structure-aware dimensionality reduction that improves structural fidelity without sacrificing predictive performance.

\section{Introduction}
\label{sec:Introduction}
Multivariate datasets in fields like biology, neuroscience, high-energy physics, and finance often exhibit complex dependencies that can challenge the assumption of \emph{isotropic residual noise} \cite{tipping1999probabilistic} often invoked in classical dimensionality reduction.
Nevertheless, principal component analysis (PCA) remains a cornerstone of data exploration owing to its transparency, analytic tractability, and direct interpretability \cite{jolliffe1986principal}. PCA yields orthogonal components that maximize variance and establish a linear correspondence between observed variables and principal directions.  However, in the presence of structured dependencies arising from spatial or temporal couplings, PCA can conflate signal and noise, thereby reducing
interpretability \cite{dyer2023why}. A variety of extensions have been proposed to mitigate these limitations: manifold-learning approaches such as diffusion maps \cite{coifman2005diffusion} and Laplacian or Hessian eigenmaps \cite{donoho2003hessian,giannakis2012nlsa} that recover smooth, low-dimensional geometry underlying high-dimensional data distributions; structured spiked-matrix models that characterize inference limits under correlated noise \cite{barbier2023fundamental}; graph- and manifold-regularized variants of PCA that augment the low-rank reconstruction with a Laplacian smoothness penalty \cite{jiang2013graph,zhang2013lowrank,shahid2015robust,feng2019pca}; and formulations that reinterpret PCA within information-geometric or dynamical frameworks \cite{shinn2023phantom,quinn2019inpca}. Together, these developments have reframed PCA as a general platform for structured and interpretable representation learning.

Here we introduce \emph{Graph Regularized PCA (GR-PCA)}, a graph-based principal component analysis method that explicitly integrates conditional dependency structure into the estimation of principal components. The method first estimates a precision matrix to infer the feature graph, then combines a sparsification step with a Laplacian regularization to bias the loadings toward sign-coherent, smooth patterns on the graph induced by the non-zero entries of the precision estimate. The sparsification promotes interpretability by selecting a compact set of features that contribute most strongly to each component.
At the same time, the Laplacian term suppresses high-frequency signals (noise) by identifying them with high-frequency Fourier modes (i.e., Laplacian eigenvectors with large eigenvalues) \cite{HAMMOND2011129}, thereby preserving coherent low-frequency modes consistent with the underlying conditional dependencies.
The resulting components are both interpretable and robust, as they capture meaningful low-frequency signals when high-frequency ones are strong, yet reduce to classical PCA when high-frequency modes are rotationally invariant. In doing so, GR-PCA unifies sparsity, smoothness, and structure awareness within a single optimization framework, providing a simple, scalable, and theoretically grounded route to inference that leverages both signal and noise structure. 

\section{Methodology}
\label{sec:Methodology}
Let $\mathbf{X} \in \mathbb{R}^{n \times p}$ denote a data matrix whose columns are centered and standardized. Given a target rank $r \in \mathbb{N}$ that controls the intrinsic dimensionality of the approximation (i.e., the number of principal components), we seek score and loading matrices $\mathbf{U} \in \mathbb{R}^{n \times r}$ and $\mathbf{V} \in \mathbb{R}^{p \times r}$, respectively, such that $\|\mathbf{X} - \mathbf{U}\mathbf{V}^\top\|_{M} \leq \epsilon$, where $\|\cdot\|_{M}$ denotes a matrix norm (e.g., the Frobenius norm), and $\operatorname{rank}(\mathbf{U}\mathbf{V}^\top) \le r$. To capture conditional dependencies among features, we estimate a sparse precision matrix $\widehat{\mathbf{\Theta}} \in \mathbb{R}^{p \times p}$ from $\mathbf{X}$ and construct an undirected graph with adjacency matrix $\mathbf{A}_{ij} = \mathbbm{1}\{|\widehat{\mathbf{\Theta}}_{ij}| > 0\}$, zero diagonal, node degrees $d_j = \sum_k \mathbf{A}_{jk}$, degree matrix $\mathbf{D} = \mathrm{diag}(d)$, and combinatorial Laplacian $\mathbf{L} = \mathbf{D} - \mathbf{A}$. Intuitively, $\mathbf{A}$ encodes which variables are directly connected and thus conditionally dependent, $\mathbf{D}$ quantifies the total connection strength of each variable, and $\mathbf{L}$ acts as the infinitesimal generator of a diffusion process on the graph, providing a natural measure of signal smoothness\,\footnote{A signal can be any vector in $\mathbb{R}^{p}$ that assigns a value to each node.}. This relationship becomes evident by considering the quadratic form of the Laplacian and the loading vectors $\mathbf{v}_k \in \mathbb{R}^p$, where $\mathbf{v}_k$ denotes the $k$-th column of $\mathbf{V}$ (i.e., principal component $k$). The quantity $\mathbf{v}_k^\top \mathbf{L} \mathbf{v}_k = \tfrac{1}{2} \sum_{i,j} \mathbf{A}_{ij} (\mathbf{v}_{k,i} - \mathbf{v}_{k,j})^2$ measures the total variation of the $k$-th loading vector across connected nodes. For each principal component, the Laplacian term accumulates the squared differences in loadings between every pair of connected features $(i,j)$ in the graph: edges connecting nodes with similar loadings contribute little, whereas edges linking nodes with very different loadings contribute strongly. In particular, when two connected coefficients have opposite signs, their difference $(\mathbf{v}_{k,i} - \mathbf{v}_{k,j})$ is large in magnitude, producing a strong penalty. Extending this to all components, we sum over columns of $\mathbf{V}$ and use the trace property

\begin{equation}
\frac12 \sum_{k=1}^r \sum_{i,j} \mathbf{A}_{ij} \,(\mathbf{v}_{k,i} - \mathbf{v}_{k,j})^2 \;=\; \mathrm{tr}\,(\mathbf{V}^\top \mathbf{L} \mathbf{V}).
\end{equation}

The link between the characteristic graph structure and signal smoothness is formally established by \cite{k4bx-w273}. Low-frequency signals vary slowly across connected nodes and thus capture large-scale, or coarse, structures of the graph \cite{7552590}. They correspond to slow diffusion dynamics. Such modes are typically dominated by nodes with relatively high degree, reflecting their central role in the overall connectivity pattern. Together, these observations motivate the following objective function, which integrates low-rank reconstruction, smoothness of the loading vectors over the graph, and sparsity:

\begin{equation}
\label{eq:ggspca:objective}
\min_{U,V}\;\;
\frac{1}{2}\,\|\mathbf{X}-\mathbf{U}\mathbf{V}^\top\|_F^2
\;+\;\alpha \sum_{j=1}^p  \,\frac{\|\mathbf{V}_{j:}\|_1}{1 + d_j}
\;+\;\frac{\lambda}{2}\,\mathrm{tr}\,\!\big(\mathbf{V}^\top \mathbf{L} \mathbf{V}\big),
\quad \alpha>0,\;\lambda\ge 0.
\end{equation}

The \emph{first term} measures the reconstruction error between $\mathbf{X}$ and its rank-$r$ approximation $\mathbf{U}\mathbf{V}^\top$. The \emph{second term} imposes sparsity on the vector of loadings associated with feature \( j \) across all principal components $\mathbf{V}_{j:}\in\mathbb{R}^{1\times r}$ through a weighted \( \ell_1 \) penalty. The weights \( (1 + d_j)^{-1} \) modulate the strength of the regularization, penalizing weakly connected features more strongly than highly connected ones. This design promotes interpretable loadings that remain consistent with the underlying graph topology. The \emph{third term}, $\frac{\lambda}{2}\,\mathrm{tr}\,(\mathbf{V}^\top \mathbf{L} \mathbf{V})$, enforces graph smoothness by encouraging connected features to have loadings of similar magnitude and sign, thereby forming sign-coherent regions over the graph \cite{7552590}. When $\lambda=0$, the graph structure is ignored. As $\lambda$ increases, the loadings become progressively smoother over the graph, with $\lambda \to \infty$ resulting in a constant loading over the graph. Together, the sparsity and smoothness terms force the solution of Eq.~\ref{eq:ggspca:objective} to reflect the connectivity structure encoded in $\boldsymbol{\widehat{\Theta}}$, while maintaining loadings sparse. The two coefficients are interpretable controls rather than performance knobs: $\alpha$ controls the support size of the loadings and $\lambda$ their smoothness.

We make three methodological choices in the graph construction. \textit{First}, when the number of features approaches or exceeds the number of samples, the sample covariance is rank deficient and its inverse is numerically unstable. Consequently, $\widehat{\mathbf{\Theta}}$ must be produced by a regularized estimator. Thus, we use the graphical lasso. However, any consistent sparse estimator of the conditional-independence structure \cite{meinshausen2006high, barfuss2016parsimonious, massara2019learning} can be substituted. \textit{Second}, we retain only the support of $\widehat{\mathbf{\Theta}}$ and discard the signs of its entries. We do this because the penalty encodes a smoothness prior on the signal loadings, namely, that directly dependent features carry similar loadings. This is an assumption about the structure of the signal rather than about the sign of a partial correlation, which reflects the conditional dependence structure of the residual noise. The signs of the loadings themselves are left free and are fixed by the data-fidelity and sparsity terms. A signed Laplacian, which penalizes the sum of loadings across negatively weighted edges, provides an extension for settings where signed partial correlations can be estimated reliably. \textit{Third}, for transparency, we use the unweighted graph, so that the penalty depends only on whether two features are conditionally dependent and not on the magnitude of their partial correlation. Spectral sparsifiers that preserve the Laplacian through effective resistances \cite{spielman2011graph,kelner2020learning} offer a weighted extension when fidelity of network dynamics is the priority \cite{mercier2022effective}.
\section{Data}
\label{sec:Data}
We construct a controllable synthetic data generator that mirrors the modeling assumptions in Eq.~\eqref{eq:ggspca:objective}: loadings are \emph{sparse}, \emph{sign--coherent}, and \emph{smooth on a feature graph}, while strong but undesirable high-frequency components are \emph{sparse yet non-smooth}. This is the setting in which GR-PCA is expected to outperform PCA because the high-frequency nuisance carries variance comparable to the signal and the leading sample eigenvectors mix the two. Thus, PCA returns directions that blend signal and noise, whereas the smoothness prior recovers the signal subspace. The corresponding recovery threshold for closely related structured spiked-matrix models is characterized in \cite{barbier2023fundamental}. Whether this advantage is realized in practice depends on how accurately the underlying feature graph can be recovered. We study two regimes: (i) an \emph{oracle} setting in which training receives the true precision graph, and (ii) a \emph{plug-in} setting in which the graph is estimated from data. In either setting, the synthetic construction is deliberate and ensures that the true signal and nuisance subspaces are known exactly, so that their recovery can be evaluated against a known ground truth rather than inferred from observational data.

\paragraph{Step 1: Definition of the features' structure} We first specify the structure of the $p$ features by defining an undirected graph $G = (V, E)$ with node set $V = \{1, \dots, p\}$ and arbitrary edge set $E \subseteq V \times V$. Together with its Laplacian $\mathbf{L}$, this graph provides the structural scaffold: it defines the adjacency underlying low-frequency components, it identifies low-degree nodes for high-frequency ones, and it determines the conditional dependency pattern shared among both the signal and the noise structure.

\paragraph{Step 2: Generation of true loadings} Given the feature graph $G=(V,E)$, we generate the \emph{ground-truth loading matrix} $\mathbf{V}_\star \in \mathbb{R}^{p\times r}$. For each component $k=1,\dots,r$, we select a center node $c_k \in V$ and a radius $\rho_k$ (i.e., neighbors, neighbors of neighbors, \dots) defining its spatial support. Let $b_k \in \mathbb{R}^p$ the hard mask with $(b_k)_j = 1$ if $\operatorname{dist}_G(j,c_k) \le \rho_k$ and $0$ otherwise. Each principal component is obtained by graph-smoothing $b_k$ through a Tikhonov filter \cite{shuman2013emerging}, followed by soft-thresholding $S_\omega$, and $\ell_2$ normalization

\begin{equation}
\mathbf{v}_k \;=\; \frac{S_\omega\,\!\big((\mathbf{I}_p + \gamma \mathbf{L})^{-1} b_k\big)}{\big\| S_\omega\,\!\big((\mathbf{I}_p + \gamma \mathbf{L})^{-1} b_k\big) \big\|_2},
\end{equation}

\noindent where $\mathbf{I}_p$ is the $(p \times p)$ identity matrix, and $\gamma>0$ controls the smoothness.

\paragraph{Step 3: Introduction of nuisance loadings} To challenge recovery of the true low-frequency structure, we add $q$ sparse high-frequency components (nuisance components). Let the sub-maximal degree node set be $\partial G = \{ j \in V : d_j < \max_{i \in V} d_i \}$; for each $\ell = 1,\dots,q$, we sample a small subset $Q_\ell \subset \partial G$ of size $|Q_\ell| \ll p$ uniformly at random and assign random signs to form

\begin{equation}
(\tilde{\mathbf{v}}_\ell^{(\nu)})_j =
\begin{cases}
\pm 1, & j \in Q_\ell, \\[3pt]
0, & \text{otherwise},
\end{cases}
\quad
\mathbf{v}_\ell^{(\nu)} = \frac{\tilde{\mathbf{v}}_\ell^{(\nu)}}{\|\tilde{\mathbf{v}}_\ell^{(\nu)}\|_2}, 
\end{equation}

\noindent where the signs of the active entries $j \in Q_\ell$ are chosen independently with equal probability (i.e., a Rademacher distribution). Collecting columns gives $\mathbf{V}_\nu = [\mathbf{v}^{(\nu)}_1,\dots,\mathbf{v}^{(\nu)}_q] \in \mathbb{R}^{p\times q}$. By design, these principal components
lie on weakly connected boundary regions, have large Laplacian quadratic form $(\mathbf{v}^{(\nu)}_\ell)^\top \mathbf{L}\,\mathbf{v}^{(\nu)}_\ell$, and lack sign coherence, producing variability associated with high-frequency graph patterns that compete with the true, smooth components.

\begin{figure}[h!]
  \centering
  \includegraphics[width=\linewidth]{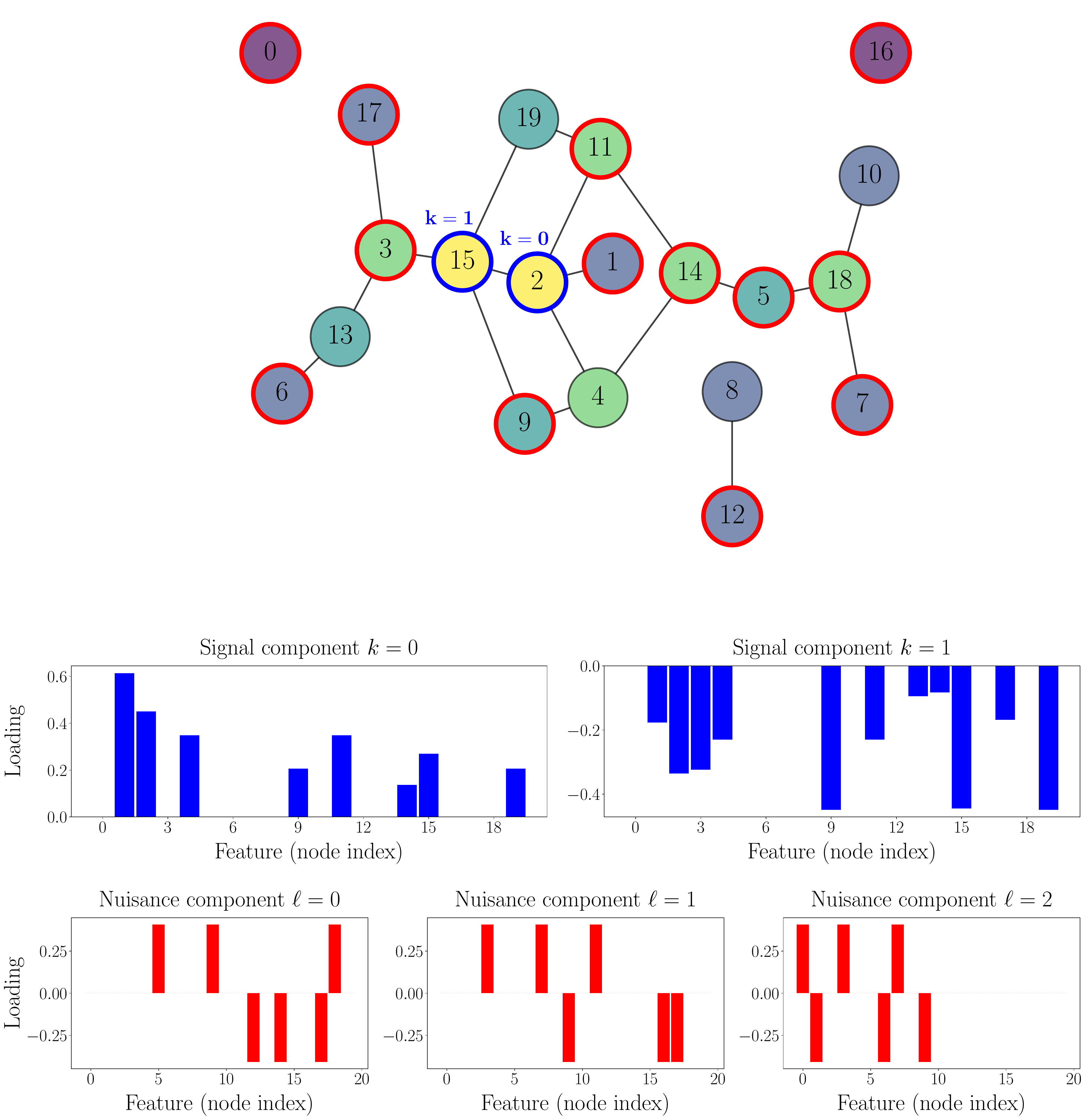}
  \caption{
  Illustration of the the first three steps of the data generation pipeline. The feature structure is given by an Erd\H{o}s--R\'enyi graph with $|V| = 20$ and edge probability $0.12$. The colormap represents node degree, shifting from purple (lowest) to yellow (highest). \textbf{(Top Histogram Panels)} Two smooth signal components ($r = 2$) generated via Tikhonov smoothing ($\gamma = 3$) and soft-thresholding ($S_\omega = 0.4$). Observe that the loadings are sign-coherent and their magnitude decays smoothly with distance from the central seed node, respecting the graph topology. \textbf{(Bottom Histogram Panels)} Three nuisance components ($q = 3$), each consisting of $s = 6$ randomly signed spikes on sub-maximal degree nodes. In contrast to the signal, these loadings are spatially incoherent and visually disjoint from the underlying community structure.}
  \label{fig:er_all_in_one}
\end{figure}

\paragraph{Step 4: Sampling of 
scores}
Given the loading matrices $\mathbf{V}_\star$ and $\mathbf{V}_\nu$, we draw sample-specific coefficients determining how strongly each observation expresses each principal component. Let $\mathbf{U}_\star \in \mathbb{R}^{n\times r}$ and $\mathbf{U}_\nu \in \mathbb{R}^{n\times q}$ be the score matrices for true and nuisance components, respectively. We generate their columns independently as

\begin{equation}
    \mathbf{U}_\star^{(:,k)} \sim \mathcal{N}(\mathbf{0},\sigma_k^2 \mathbf{I}_n), \qquad
    \mathbf{U}_\nu^{(:,\ell)} \sim \mathcal{N}(\mathbf{0},\sigma_\ell^2 \mathbf{I}_n),
\end{equation}

\noindent where $\mathbf{I}_n$ denotes the $(n \times n)$ identity matrix, ensuring that observations are independent and identically distributed. The variances $\sigma_k^2$ and $\sigma_\ell^2$ represent the component strengths; we set them to decay geometrically (e.g., $\sigma_k^2=\sigma_1^2\rho^{k-1}$ with $\rho<1$) so that early components dominate. These mean-zero Gaussian draws yield realistic random variation without introducing structured correlations between the components themselves.

\paragraph{Step 5: Injection of graph-correlated noise}

To model graph-consistent measurement error, we add Gaussian noise with precision

\begin{equation}
    \label{eq:noise_precision_matrix}
    \boldsymbol{\Theta} = \tau \mathbf{I}_p + \beta \mathbf{L}, \qquad 
    \varepsilon_i \sim \mathcal{N}(\mathbf{0},\boldsymbol{\Theta}^{-1}), \; i=1,\dots,n,
\end{equation}

\noindent where $\tau>0$ controls the independent noise precision, and $\beta\ge0$ sets neighbor coupling. This yields locally correlated noise aligned with the feature graph, embedding the same structure in both signal and noise.

\paragraph{Step 6: Generation of data samples} 

Synthetic data are finally generated by adding smooth, nuisance, and noise components:

\begin{equation}
    X = \mathbf{U}_\star \mathbf{V}_\star^\top + \mathbf{U}_\nu \mathbf{V}_\nu^\top + \sigma_E \mathbf{E},
\end{equation}

\noindent where $\mathbf{E}$ is the graph-correlated noise, and $\sigma_E>0$ sets its scale.  
Each feature column of $\mathbf{X}$ is then standardized to zero mean and unit variance, yielding data whose principal component structure, nuisance variability, and noise are fully consistent with the feature graph.

\section{Evaluation Metrics}
\label{sec:Evaluation_Metrics}

The effectiveness of GR-PCA is evaluated on synthetic datasets (see Sec.~\ref{sec:Data}) and compared against the two well-established baseline methods, PCA and SparsePCA~\cite{zou2006sparse}. All models are trained on the training data and assessed on held-out test sets using a 5-fold cross-validation procedure. Performance is measured using three evaluation metrics, and for all of them higher values correspond to better performance.

\paragraph{Selectivity score}%
Let $\mathbf{V_\star} \in \mathbb{R}^{p\times r_\star}$ and $\mathbf{V_\nu} \in \mathbb{R}^{p\times r_\nu}$ denote the ground-truth loadings for the ``true'' (smooth, sign-coherent) and ``nuisance'' (non-smooth) subspaces, each scaled consistently with the test data. 
Given the centered and standardized test matrix $\mathbf{X}_{\mathrm{te}}$ and its reconstruction $\widehat{\mathbf{X}}_{\mathrm{te}}$, 
we project both onto the corresponding subspaces

\begin{equation}
    \Pi_\star = \mathbf{V_\star}(\mathbf{V_\star}^\top \mathbf{V_\star})^{-1}\mathbf{V_\star}^\top,
    \qquad
    \Pi_\nu   = \mathbf{V_\nu}  (\mathbf{V_\nu}^\top  \mathbf{V_\nu})^{-1}\mathbf{V_\nu}^\top .
\end{equation}

\noindent The fraction of variance explained within each subspace is:

\begin{equation}
    R^2_{\mathrm{true}}
    = 1 - \frac{\lVert \mathbf{X}_{\mathrm{te}}\Pi_\star - \widehat{\mathbf{X}}_{\mathrm{te}}\Pi_\star\rVert_F^2}
            {\lVert \mathbf{X}_{\mathrm{te}}\Pi_\star\rVert_F^2},
    \qquad
    R^2_{\mathrm{nuis}}
    = 1 - \frac{\lVert \mathbf{X}_{\mathrm{te}}\Pi_\nu - \widehat{\mathbf{X}}_{\mathrm{te}}\Pi_\nu\rVert_F^2}
            {\lVert \mathbf{X}_{\mathrm{te}}\Pi_\nu\rVert_F^2}.
\end{equation}

\noindent The \emph{selectivity score} ($\Delta$) is their difference,

\begin{equation}
    \Delta
    = R^2_{\mathrm{true}} - R^2_{\mathrm{nuis}},
    \quad
    \Delta\in[-1,1],
\end{equation}

\noindent which increases when the model captures variance along the true subspace while suppressing variance along nuisance directions. 

\paragraph{Alignment score} Let $\mathbf{V_\star} = [v^\star_1,\dots,v^\star_{r_\star}] \in \mathbb{R}^{p\times r_\star}$ be the ground-truth loadings and 
$\widehat{\mathbf{V}} = [\widehat v_1,\dots,\widehat v_{\widehat{r}}] \in \mathbb{R}^{p\times \widehat{r}}$ their estimated counterparts, 
each column normalized to have unit length. 
To compare the two sets of components, we measure how well each true direction $v^\star_k$ aligns with every estimated one using the absolute cosine similarity $\big|\langle v^\star_k,\, \widehat v_j \rangle\big|$. We then find the pairing between true and estimated components that gives the \emph{largest total similarity}, 
ensuring each component is used at most once. 
If one set has extra components, the unmatched ones are treated as having similarity zero. 
The \emph{alignment score} is the mean similarity across the matched pairs:

\begin{equation}
    \mathrm{align}(\widehat{\mathbf{V}}, \mathbf{V_\star})
    = \frac{1}{r_\mathrm{match}}
    \sum_{(k,j)\in \mathcal{M}^\star}
    \big|\langle v^\star_k,\, \widehat v_j\rangle\big|,
    \quad
    \mathrm{align}\in[0,1],
\end{equation}

\noindent where $\mathcal{M}^\star$ denotes the optimal one-to-one matching and $r_\mathrm{match}=\min(r_\star,\hat{r})$. 
A score of $1$ indicates perfect correspondence, 
while a score near $0$ means the estimated directions are nearly orthogonal to the truth. Each matched term is the cosine of the angle between the true direction and an estimated direction. It is therefore in monotone correspondence with the squared-sine error $\sin^2\theta = 1 - \cos^2\theta$, which is standard in the PCA literature \cite{huang2021streaming}. 

\paragraph{Global reconstruction score}
Given the centered and standardized test matrix $\mathbf{X}_{\mathrm{te}}$ and its reconstruction $\widehat{\mathbf{X}}_{\mathrm{te}}$, the overall fidelity of reconstruction is measured by the \emph{global coefficient of determination}:

\begin{equation}
    R^2_X
    = 1 - \frac{\lVert \mathbf{X}_{\mathrm{te}} - \widehat{\mathbf{X}}_{\mathrm{te}}\rVert_F^2}
            {\lVert \mathbf{X}_{\mathrm{te}}\rVert_F^2},
    \quad
    R^2_X\in[0,1].
\end{equation}

\noindent Although $R^2_X = 1$ indicates perfect reconstruction, we treat it as a control metric rather than a primary measure of methodological success. Effective recovery of low-frequency structure combined with suppression of high-frequency variability will necessarily reduce $R^2_X$ below 1. Critically, such reduction is not inherently diagnostic: in the presence of anisotropic noise, baseline methods (i.e., PCA and SparsePCA) may also exhibit low fidelity due to their failure to model noise structure rather than its deliberate suppression. A correct interpretation of $R^2_X$ requires joint consideration with the selectivity and alignment metrics introduced earlier in this section, which together distinguish guided suppression of nuisance variation from unstructured reconstruction error.

\section{Results}
\label{sec:Results}

To assess the ability of each estimator (i.e., PCA, SPCA, GR-PCA) to concentrate variance within the low-frequency subspace, we evaluate selectivity under controlled regimes of isotropic and anisotropic graph-correlated noise. We employ three matched-sparsity\footnote{The term ``matched-sparsity'' indicates that the Erd\H{o}s--R\'enyi (ER), Barab\'asi--Albert (BA), and Watts--Strogatz (WS) graphs are constructed to have comparable, though not identical, edge densities. Exact equivalence is nontrivial because each graph topology imposes structural constraints (e.g., fixed expected degree in ER, minimum degree in BA, and fixed neighborhood size in WS) that restrict the set of feasible densities.} feature structures: Erd\H{o}s--R\'enyi (ER), Barab\'asi--Albert (BA), and Watts--Strogatz (WS) \cite{barabasi2013network}. While not exhaustive, this selection spans distinct topological regimes, serving as prototypical examples of degree homogeneity, scale-free heterogeneity, and small-world clustering, respectively. The three graph families span the relevant range of spectra associated with graphs whose degree distributions range from homogeneous to heavy-tailed.

\begin{table}[h!]
\centering
\caption{Parameters of the synthetic data generator for the isotropic and anisotropic settings.}
\label{tab:parametrization}
\begin{tabular}{l|c|cc}
\toprule
\textbf{Parameter} & \textbf{Symbol} & \textbf{Isotropic} & \textbf{Anisotropic} \\
\midrule
Features & $p$ & $144$ & $144$ \\
Samples & $n$ & $10{,}000$ & $10{,}000$ \\
Low-frequency components & $r$ & $8$ & $8$ \\
Smoothing & $\gamma$ & $16.0$ & $16.0$ \\
Soft-threshold & $S_{\omega}$ & $0.4$ & $0.4$ \\
Nuisance spikes & $s$ & $60$ & $60$ \\
\midrule
Nuisance-to-signal ratio & $q$ & $0.1$ & $2$ \\
Independent noise precision & $\tau$ & $0.55$ & $0.10$ \\
Neighbor coupling & $\beta$ & $1.15$ & $2.50$ \\
Noise scale & $\sigma_{E}$ & $1.0$ & $3.0$ \\
\bottomrule
\end{tabular}
\end{table}

The distinction between isotropic and anisotropic variability arises from differences in the structure of the noise precision matrix (see Eq. \eqref{eq:noise_precision_matrix}) and in the relative magnitude and spatial distribution of nuisance variability across the graph. In the isotropic regime, the diagonal term is comparatively large and the nuisance load is reduced, so graph-correlated noise remains homogeneous across features. In the anisotropic regime, the diagonal term is weak and the graph term is larger; at the same time we inject more nuisance components at larger amplitude, making the marginal variance of a feature depend sharply on its degree and local clustering. The two regimes are parametrized as reported in Table \ref{tab:parametrization}.

Under this specification, the isotropic regime produces nearly rotationally invariant noise once features are standardized. The anisotropic regime, by contrast, produces graph-structured noise via $\mathbf{L}$, boosting the nuisance-to-signal ratio and the overall variability. In all the experiments, we report results for both GR-PCA (oracle), which uses the precision matrix from the generator, and GR-PCA (learned), in which the precision matrix is estimated from the data by GraphicalLassoCV. The comparison between using the true and estimated precision graphs isolates how graph-estimation error propagates to the recovered components. We work in the regime $n \gg p$ so that this gap is not dominated by finite-sample limitations of the precision estimator. Nevertheless, as the density of edges increases in the following experiments, estimation of the precision graph eventually becomes unreliable.

\begin{figure}[h!]
    \centering
    \begin{subfigure}[t]{0.48\textwidth}
        \centering
        \includegraphics[width=\linewidth]{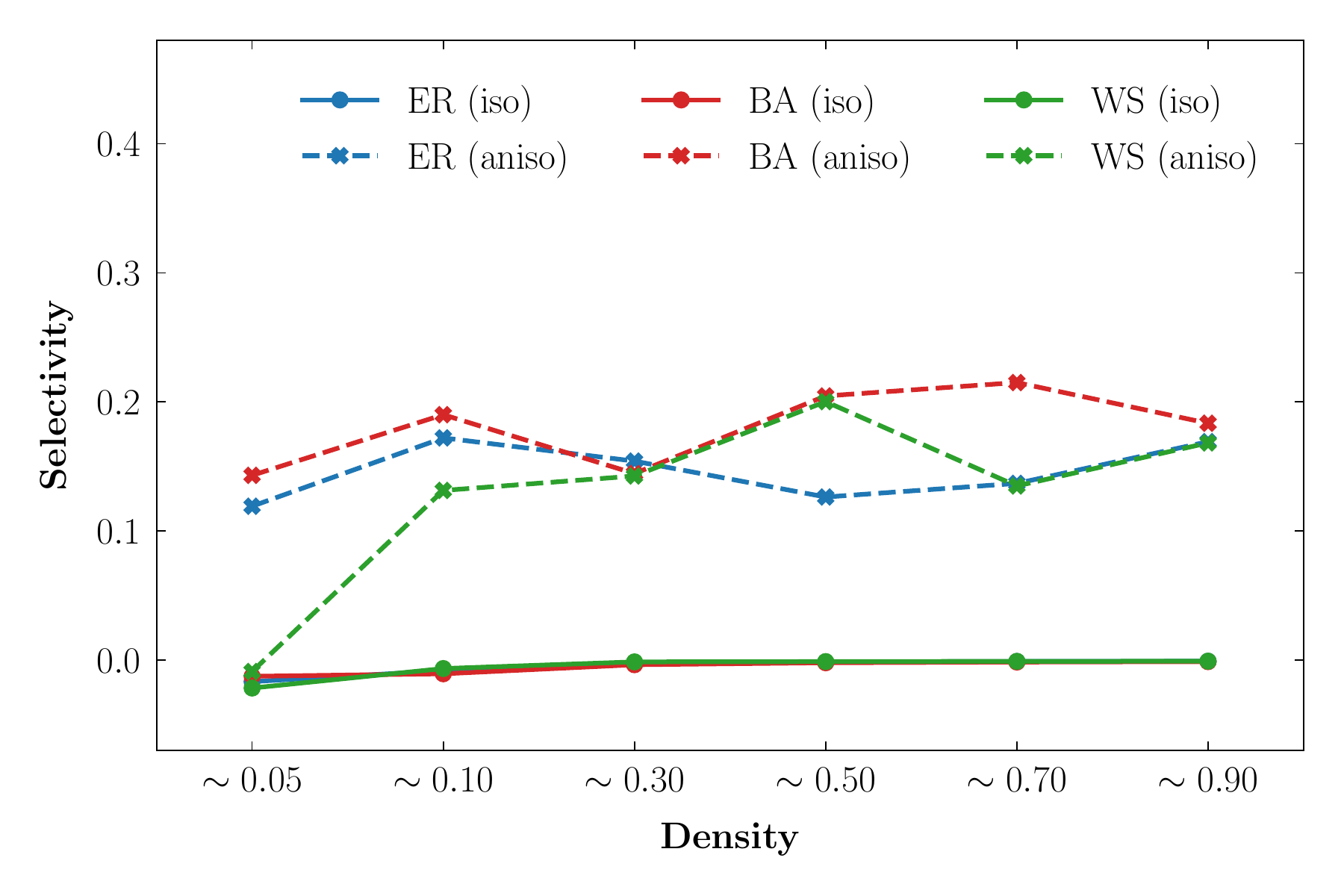}
        \caption{\textbf{PCA}}
        \label{fig:pca_density}
    \end{subfigure}
    \hfill
    \begin{subfigure}[t]{0.48\textwidth}
        \centering
        \includegraphics[width=\linewidth]{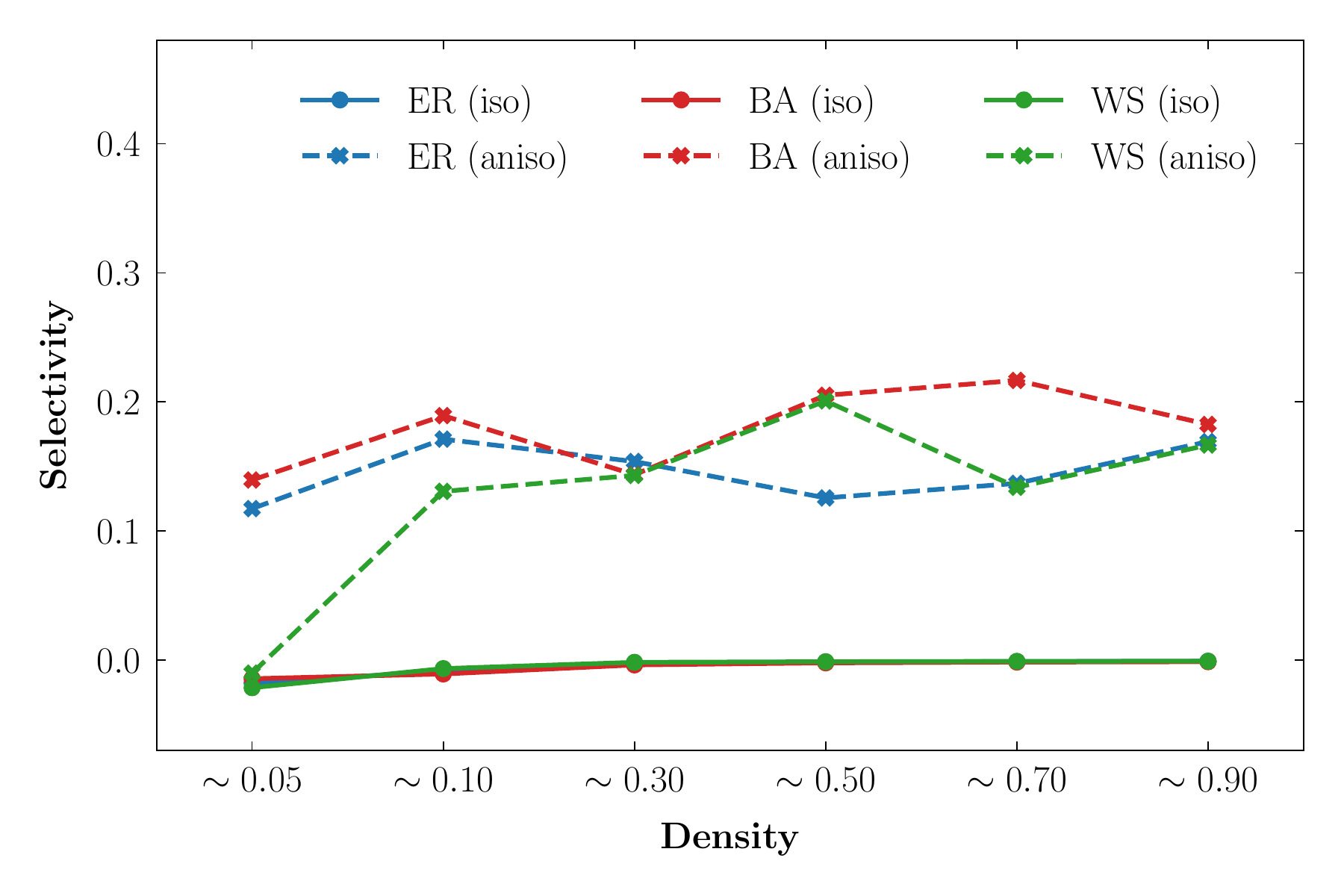}
        \caption{\textbf{SparsePCA}}
        \label{fig:spca_density}
    \end{subfigure}

    \vspace{0.5em}

    \begin{subfigure}[t]{0.48\textwidth}
        \centering
        \includegraphics[width=\linewidth]{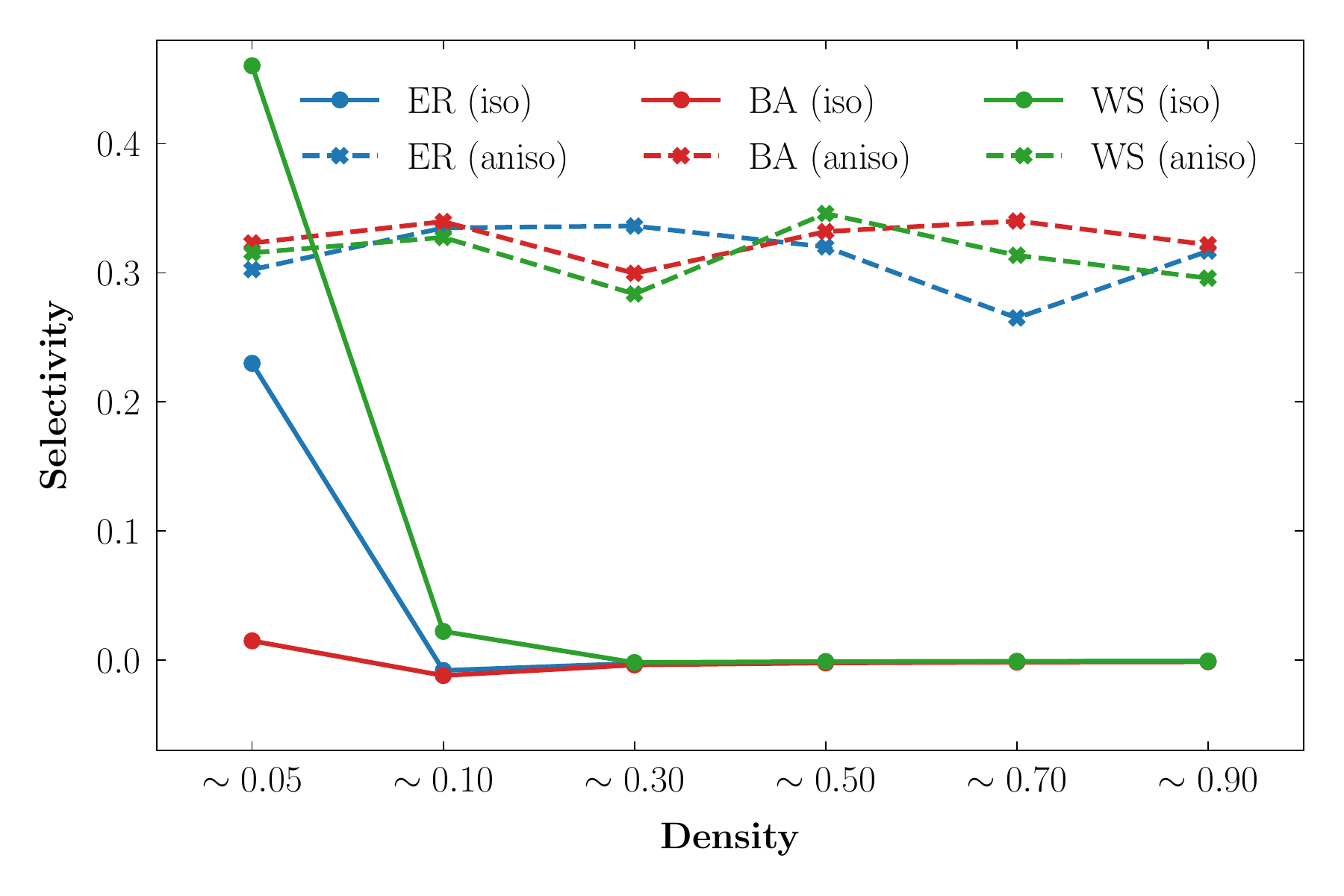}
        \caption{\textbf{GR-PCA (oracle)}}
        \label{fig:grpca_oracle_density}
    \end{subfigure}
    \hfill
    \begin{subfigure}[t]{0.48\textwidth}
        \centering
        \includegraphics[width=\linewidth]{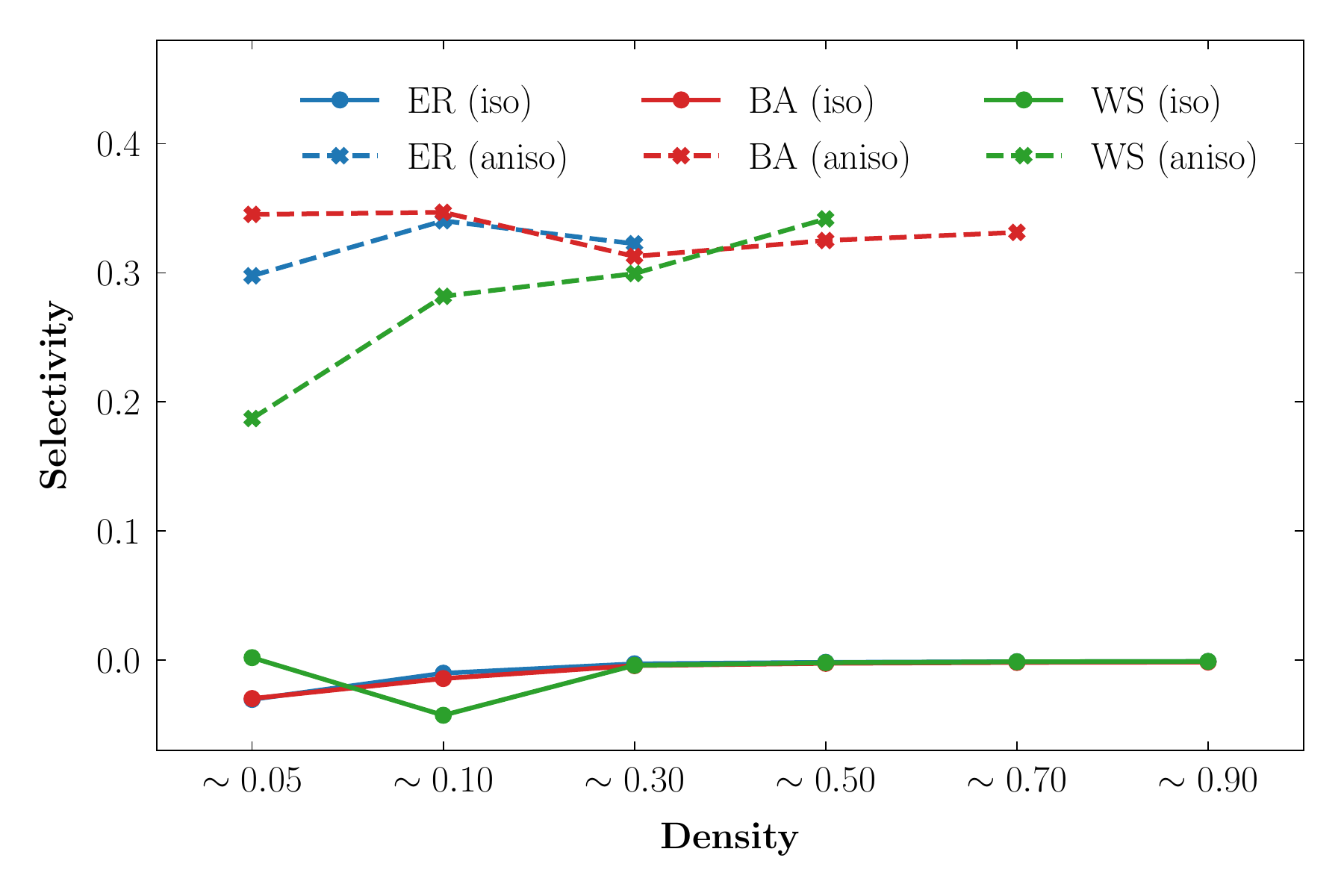}
        \caption{\textbf{GR-PCA (learned)}}
        \label{fig:grpca_learned_density}
    \end{subfigure}

    \caption{
        Selectivity as a function of edge density for the four PCA variants, evaluated across the three feature structures (ER in blue, BA in red, WS in green). Solid lines correspond to the isotropic noise regime, while dashed lines indicate the anisotropic regime.
    }
    \label{fig:selectivity_vs_density_all}
\end{figure}

Across all analyses, we use the same organizing principle. If a method concentrates variance on the low-frequency subspace and suppresses high-frequency components, we expect three linked outcomes: (\emph{i}) increased selectivity because variance explained shifts toward low-frequency directions and away from high-frequency ones, (\emph{ii}) higher alignment because estimated directions coincide with the true low-frequency components, and (\emph{iii}) a controlled reduction in global \(\,R_X^2\,\) in anisotropic settings because some high-frequency variance is intentionally discarded.

\figurename~\ref{fig:selectivity_vs_density_all} summarizes the \emph{selectivity} performance as a function of density, where each curve tracks the matched-sparsity ER, BA, and WS feature structures under isotropic (solid) and anisotropic (dashed) regimes. Standard PCA and SparsePCA exhibit near-zero selectivity in the isotropic regime (approximately $-0.005$ across all feature structures) and achieve values between $0.128$ and $0.180$ in the anisotropic regime. The narrow range across feature structures (no larger than $0.052$) confirms the initial hypothesis that variance maximization alone does not adapt to structural dependencies in the feature graph and therefore does not prioritize recovery of low-frequency components over high-frequency components.

In the anisotropic regime, graph-regularized PCA achieves substantially higher selectivity than unregularized baselines, and the magnitude of this improvement depends both on the feature structure and on whether the precision graph is supplied or learned. On ER and BA feature structures, the oracle variant yields mean selectivity values of $0.313$ and $0.326$, respectively, while the learned variant achieves $0.320$ and $0.332$, respectively. These 
differences ($\leq 0.009$) indicate that the precision graph inferred from data preserves the information needed to enhance the recovery of low-frequency components and suppress high-frequency ones. In these settings, estimating the precision graph may remove edges that primarily support high-frequency components, strengthening the preference for low-frequency structure and enabling the learned variant to outperform the oracle variant slightly.

\begin{table}[h!]
\centering
\caption{Mean out-of-sample selectivity (averaged over all graph densities) across regimes and feature structures.}
\label{tab:selectivity_topologies}
\begin{tabular}{lcccccc}
\toprule
& \multicolumn{3}{c}{\textbf{Isotropic}} &
  \multicolumn{3}{c}{\textbf{Anisotropic}} \\
\cmidrule(lr){2-4} \cmidrule(lr){5-7}
\textbf{Method} & ER & BA & WS & ER & BA & WS \\
\midrule
PCA              & $-0.005$ & $-0.005$ & $-0.005$ & $0.146$ & $0.180$ & $0.128$ \\
SparsePCA        & $-0.005$ & $-0.006$ & $-0.005$ & $0.146$ & $0.180$ & $0.128$ \\
\midrule
GR-PCA (oracle)  & $\;\;\;\underline{\mathbf{0.036}}$ & $\underline{\mathbf{-0.001}}$ & $\;\;\;\underline{\mathbf{0.080}}$ & $0.313$ & $0.326$ & $\underline{\mathbf{0.314}}$ \\
GR-PCA (learned) & $-0.008$ & $-0.009$ & $-0.008$ & $\underline{\mathbf{0.320}}$ & $\underline{\mathbf{0.332}}$ & $0.278$ \\
\bottomrule
\end{tabular}
\end{table}

In the case of WS feature structure, the oracle variant reaches a mean selectivity of $0.314$, whereas the learned variant drops to $0.278$, a gap of $0.036$ under identical experimental conditions. Two mechanisms contribute to this reduced performance. First, WS graphs contain densely clustered neighborhoods and short cycles, which inherently limit the separation between low- and high-frequency components. Second, this limited separation makes precision graph estimation more difficult, especially at high densities, where the sample covariance matrix becomes increasingly ill-conditioned and GraphicalLassoCV struggles to converge or returns an overly dense precision graph. Related failures also occur at densities greater than $0.70$ for ER and $0.90$ for BA. In those settings, the feature graph approaches a nearly uniform structure in which conditional dependencies are no longer well distinguishable from marginal correlations. This reduces the sparsity of the true precision matrix and pushes the estimation problem toward a nearly unidentifiable regime, thereby affecting the model's ability to suppress high-frequency components.

In summary, the selectivity results establish that graph regularization, when supported by a reliably estimated precision graph, reallocates explained variance toward low-frequency components and away from high-frequency ones, with topology- and density-dependent limits. 

We next evaluate \emph{alignment} to verify that the variance reallocation documented by selectivity coincides with the recovery of the ground-truth low-frequency directions. Averaged over all settings, PCA and SparsePCA reach mean alignment scores of $0.210$ and $0.323$, respectively, indicating that a substantial portion of the variance they capture remains associated with high-frequency components. The oracle form of graph-regularized PCA achieves a mean of $0.958$, while the learned form achieves $0.485$. This overall mean for the learned model combines two distinct regimes, with low alignment in the isotropic regime and high alignment in the anisotropic regime, as reported in \autoref{tab:alignment_topologies}.

\begin{table}[h!]
\centering
\caption{Mean out-of-sample alignment (averaged over all graph densities) across regimes and feature structures.}
\label{tab:alignment_topologies}
\begin{tabular}{lcccccc}
\toprule
& \multicolumn{3}{c}{\textbf{Isotropic}} &
  \multicolumn{3}{c}{\textbf{Anisotropic}} \\
\cmidrule(lr){2-4} \cmidrule(lr){5-7}
\textbf{Method} & ER & BA & WS & ER & BA & WS \\
\midrule
PCA              & $0.137$ & $0.147$ & $0.181$ & $0.271$ & $0.256$ & $0.270$ \\
SparsePCA        & $0.357$ & $0.357$ & $0.437$ & $0.261$ & $0.256$ & $0.270$ \\
\midrule
GR-PCA (oracle)  & $\underline{\mathbf{0.996}}$ & $\underline{\mathbf{0.995}}$ & $\underline{\mathbf{0.972}}$ & $0.939$ & $0.927$ & $\underline{\mathbf{0.920}}$ \\
GR-PCA (learned) & $0.148$ & $0.157$ & $0.255$ & $\underline{\mathbf{0.947}}$ & $\underline{\mathbf{0.935}}$ & $0.916$ \\
\bottomrule
\end{tabular}
\end{table}

In the anisotropic regime, the learned variant aligns with the true low-frequency components at $0.947$ (ER), $0.935$ (BA), and $0.916$ (WS), closely tracking the oracle at $0.939$ (ER), $0.927$ (BA), and $0.920$ (WS). In the isotropic regime, consistently with the limited role of the feature graph when low-frequency structure is not emphasized by the noise, alignment for the learned variant drops to $0.148$–$0.255$ depending on the feature structure. By density, alignment remains high across most of the anisotropic sweep and only degrades when precision learning fails at the highest densities (for example, ER at density $\ge 0.50$, BA at density $0.90$, and WS at density $\ge 0.70$), matching the selectivity behavior in \autoref{fig:selectivity_vs_density_all}. Together with the selectivity analysis, these results confirm that the mechanism at work is the same in both metrics: an effective precision graph enables the regularizer to favor low-frequency components and suppress high-frequency ones.

If alignment reflects successful identification of low-frequency components, then a reduction in overall out-of-sample reconstruction is expected precisely where high-frequency components carry nontrivial variance. We assess global reconstruction accuracy using the out-of-sample \textit{coefficient of determination} ($R_X^2$). PCA and SparsePCA achieve high scores in the isotropic regime and moderate scores in the anisotropic regime, reflecting the absence of any mechanism to suppress high-frequency components. Graph-regularized models exhibit similar reconstruction performance in the isotropic regime and lower performance in the anisotropic regime, consistent with their goal of down-weighting high-frequency components that contribute to overall variance but do not belong to the low-frequency subspace. Exact means by regime and topology are reported in \autoref{tab:recon_topologies} and should be interpreted jointly with Tables~\ref{tab:selectivity_topologies} and~\ref{tab:alignment_topologies}. In the anisotropic setting, GR-PCA attains higher selectivity and alignment while exhibiting lower $R^2_X$, indicating that the discarded variance is predominantly associated with high-frequency components of the feature graph. Conversely, the higher $R^2_X$ achieved by PCA and SparsePCA in the same setting reflects reconstruction of noise-driven variability rather than improved recovery of the true low-frequency structure.

\begin{table}[h!]
\centering
\caption{Mean out-of-sample global reconstruction accuracy (averaged over all graph densities) across regimes and feature structures. As this quantity serves only as a control metric rather than a primary indicator of methodological success, individual values are not emphasized. Its interpretation should be made jointly with the selectivity (see Table~\ref{tab:selectivity_topologies}) and alignment (see Table~\ref{tab:alignment_topologies}) metrics.}
\label{tab:recon_topologies}
\begin{tabular}{lcccccc}
\toprule
& \multicolumn{3}{c}{\textbf{Isotropic}} &
  \multicolumn{3}{c}{\textbf{Anisotropic}} \\
\cmidrule(lr){2-4} \cmidrule(lr){5-7}
\textbf{Method} & ER & BA & WS & ER & BA & WS \\
\midrule
PCA              & $0.945$ & $0.937$ & $0.945$ & $0.625$ & $0.621$ & $0.624$ \\
SparsePCA        & $0.945$ & $0.937$ & $0.945$ & $0.625$ & $0.621$ & $0.624$ \\
\midrule
GR-PCA (oracle)  & $0.931$ & $0.931$ & $0.915$ & $0.499$ & $0.493$ & $0.495$ \\
GR-PCA (learned) & $0.940$ & $0.932$ & $0.927$ & $0.472$ & $0.472$ & $0.472$ \\
\bottomrule
\end{tabular}
\end{table}

The reconstruction gap between learned and oracle variants in the anisotropic regime is $0.021$–$0.027$ by feature structure, consistent with a stronger penalty on high-frequency components when the precision graph is known. For the learned variant, the decline in $R_X^2$ is most pronounced at densities where precision learning becomes unstable, which is the same region where selectivity and alignment also deteriorate. 

Integrating all these findings across all three evaluation metrics provides a consistent picture of how graph regularization operates. When the precision graph is available or can be learned reliably, graph-regularized PCA reallocates explained variance toward the low-frequency subspace (higher selectivity), recovers the intended directions (higher alignment), and exhibits a predictable reduction in overall reconstruction only in settings where high-frequency components carry substantial variance (lower $R_X^2$ under anisotropy). The benefits of graph regularization therefore depend critically on the presence of a meaningful feature structures: ER and BA graphs maintain a usable separation between low- and high-frequency components at most densities, enabling stable precision recovery and strong performance improvements. In contrast, when the feature graph becomes nearly fully connected at very high densities, conditional dependencies become nearly uniform, the distinction between low- and high-frequency components collapses, and graph-regularized PCA naturally converges toward the behavior of standard PCA. Within the substantial range where graph structure remains informative, the method consistently delivers structure-aware dimensionality reduction that favors relevant low-frequency components and suppresses high-frequency modes.

\section*{Conclusions}

Graph-regularized PCA (GR-PCA) demonstrates that integrating feature structure into principal component estimation enables separation between structured and unstructured variability in multivariate data. Incorporating connectivity among features reshapes the variance landscape, concentrating energy on low-frequency graph components, while pruning high-frequency ones. This leads to representations that are both interpretable and resilient to anisotropic noise, as confirmed by consistent improvements in selectivity, alignment, and stability across diverse graph topologies and noise regimes.

GR-PCA balances the variance explained and the smoothness of the component loadings across the graph, yielding principal directions that remain informative while respecting the structural organization of the features. The method provides a minimal and analytically transparent model of structure-aware inference that reveals the precise conditions under which graph information enhances recovery: a non-uniform precision structure, distinguishable low- and high-frequency components, and a stable estimation of the precision graph. Under these conditions, GR-PCA acts as a selective filter that amplifies coherent, low-frequency directions while attenuating noise-driven ones, thereby enhancing interpretability and robustness. 

Beyond its theoretical implications, GR-PCA has broad potential for applications in fields where observations are  structured \cite{briola2023topological, wang2023topological, briola2025hlob, lin2026compositional}. Examples include neuroscience \cite{maier1987principal}, where cortical activity unfolds over structural connectivity graphs; climate science \cite{demvsar2013principal}, where spatial fields are sampled on irregular grids; genomics \cite{ringner2008principal}, where gene expression follows interaction networks; and energy systems \cite{nyangon2024principal}, where nodal dynamics are coupled through transmission infrastructure.

Several open directions emerge from this foundation. A first line is the development of robust approaches for graph learning under time-varying conditions, enabling GR-PCA to operate in dynamic systems where the feature topology evolves with the process being observed. A second direction is the integration of this framework with probabilistic and nonlinear extensions of PCA, which could connect graph-based regularization to modern theories of manifold learning and uncertainty quantification.

Overall, this work presents Graph-regularized PCA as an extension of classical PCA for structured data analysis. It preserves the analytical clarity of variance-based decomposition while adding the ability to align estimation with the intrinsic organization of the features. By combining the simplicity of linear methods with the interpretive power of graph-informed regularization, GR-PCA is capable of learning directly from the geometry of complex, interconnected systems.

\printbibliography

\end{document}